\documentclass[11pt,a4paper]{article}
\usepackage[hyperref]{emnlp2020}
\usepackage{times}
\usepackage{xspace}
\usepackage{latexsym}
\usepackage{xcolor}
\usepackage{url}
\usepackage{acronym}
\usepackage{graphicx} 
\usepackage{amsmath,amssymb}
\usepackage[inline]{enumitem}
\usepackage{acronym}
\usepackage{booktabs}
\usepackage{array, makecell} 
\usepackage[skip=0pt]{caption}
\usepackage{mathtools}
\usepackage{natbib}
\usepackage{etoolbox}
\newcounter{todocnt}

\acrodef{RL}{Reinforcement Learning}
\acrodef{PL}{Policy learning}
\acrodef{DRL}{Deep Reinforcement Learning}
\acrodef{IRL}{Inverse Reinforcement Learning}
\acrodef{SERP}{search engine result page}
\acrodef{IR}{Information Retrieval}
\acrodef{MDP}{Markov Decision Process}
\acrodef{MaxEnt-IRL}{Maximum Entropy Inverse Reinforcement Learning}
\acrodef{DM-IRL}{Distance Minimization Inverse Reinforcement Learning}
\acrodef{MMI}{Maximum Mutual Information} 
\acrodef{DNN}{Deep Neural Networks}
\acrodef{RNN}{Recurrent Neural Networks}
\acrodef{MLP}{Multilayer Perceptron}
\acrodef{GRU}{Gated Recurrent Net}
\acrodef{TDS}{Task-oriented Dialogue System}
\acrodef{LU}{language understanding}
\acrodef{DM}{dialogue management}
\acrodef{NLG}{natural language generation}
\acrodef{DST}{Dialogue State Tracker}
\acrodef{DQN}{Deep Q-network}
\acrodef{PPO}{Proximal Policy Optimization}
\acrodef{GAIL}{Generative Adversarial Imitation Learning} 
\acrodef{AIRL}{Adversarial Inverse Reinforcement Learning}
\acrodef{GAN}{Generative Adversarial Network}

\aclfinalcopy 

\DeclareMathOperator*{\argmax}{arg\,max}

\title{Guided Dialogue Policy Learning without\\ Adversarial Learning in the Loop}

\author{
Ziming Li \textsuperscript{1},
Sungjin Lee \textsuperscript{2},
Baolin Peng \textsuperscript{3},
Jinchao Li  \textsuperscript{3},\\
{\bf Julia Kiseleva \textsuperscript{3},
Maarten de Rijke \textsuperscript{1},
Shahin Shayandeh \textsuperscript{3},
Jianfeng Gao \textsuperscript{3}}\\
\textsuperscript{1}University of Amsterdam,
\textsuperscript{2}Amazon,
\textsuperscript{3}Microsoft Research\\
\{z.li,m.derijke@@uva.nl\},
sungjinl@amazon.com,
\{baolin.peng,jincli,shahins,jfgao@microsoft.com\}
}

\date{}

\begin{document}
\maketitle
\begin{abstract}
\ac{RL} methods have emerged as a popular choice for training an efficient and effective dialogue policy. However, these methods suffer from sparse and unstable reward signals returned by a user simulator only when a dialogue finishes. Besides, the reward signal is manually designed by human experts, which requires domain knowledge. 
Recently, a number of adversarial learning methods have been proposed to learn the reward function together with the dialogue policy. However, to alternatively update the dialogue policy and the reward model on the fly, we are limited to policy-gradient-based algorithms, such as REINFORCE and PPO. Moreover, the alternating training of a dialogue agent and the reward model can easily get stuck in local optima or result in mode collapse. 
To overcome the listed issues, we propose to decompose the adversarial training into two steps. First, we train the discriminator with an auxiliary dialogue generator and then incorporate a derived reward model into a common \ac{RL} method to guide the dialogue policy learning. This approach is applicable to both on-policy and off-policy \ac{RL} methods.  Based on our extensive experimentation, we can conclude the proposed method: (1)~achieves a remarkable task success rate using both on-policy and off-policy \ac{RL} methods; and (2)~has potential to transfer knowledge from existing domains to a new domain.
\end{abstract}

\setlength{\abovedisplayskip}{3pt}
\setlength{\belowdisplayskip}{3pt}

\section{Introduction}
\label{sec:introduction}
\acp{TDS}, such as Siri, Google Assistant, and Amazon Alexa, aim to offer users assistance with completing tasks.
\acp{TDS} need dialogue policies to select appropriate actions at each dialogue step according to the current context of the conversation~\citep{chen2017survey}.
The development of \ac{RL} in robotics and other domains has brought a new view on learning dialogue policies~\citep{williams2007partially,gavsic2014gaussian,su2017sample}: it allows us to train with far more data than can be feasibly collected from actual users. The aim of \acp{TDS} is to maximize positive user feedback. \acp{TDS} based on \ac{RL} are amenable to training with user simulators instead of real humans~\citep{schatzmann2007agenda,li2016user}. User simulators rely on a reward function that scores system actions given dialogue context~\citep{peng2018deep,williams2017hybrid,dhingra2016towards,su2016line}.

%

The most straightforward way to design a dialogue reward function is to score the agent based on the dialogue status in a rule-based fashion: if the dialogue ends successfully, a large positive reward will be returned; if the dialogue fails, the reward will be a large negative value; if the dialogue is still ongoing, a small negative value will be returned to encourage shorter sessions~\citep{peng2018deep}. However, the rule-based solution is inflexible as it assigns the same negative reward to all the system actions before the dialogue ends. The sparse reward makes the qualities of different actions indistinguishable. Additionally, the rule-based approaches only return a meaningful reward when dialogue finishes, which can delay the penalty for low-quality actions and a high reward for high-quality ones during the conversation itself. \citet{liu2018adversarial} address the difficulties listed above by employing adversarial training for policy learning by jointly training two systems: (1)~a policy model that decides which action to take at each turn, and (2)~a discriminator that marks if a dialogue was successful or not. Feedback from the discriminator is used as a reward to push the policy model to complete a task indistinguishably from humans. Improving upon this solution, \citet{takanobu2019guided} propose to replace the discriminator with a reward function that acts at the dialogue action level and returns the reward for the given action relying on the dialogue state, system action, and next dialogue state as its input.
However, the described methods are limited to policy gradient-based algorithms, such as
REINFORCE~\citep{williams1992simple} and \ac{PPO}~\citep{schulman2017proximal}, to alternatively update the dialogue policy and the reward model on the fly, while off-policy methods are not able to benefit from self-learned reward functions. Furthermore, the alternative training of the dialogue agent and the reward model can easily get stuck in local optima or result in mode collapse. 

To alleviate the problems mentioned above, in this work we propose a new approach for training dialogue policy by decomposing the adversarial learning method into two sequential steps. First, we learn the reward function using an auxiliary dialogue state generator where the loss from the discriminator can be backpropagated to the generator directly. Second, the trained discriminator as the dialogue reward model will be incorporated into the \ac{RL} process to guide dialogue policy learning and will not be updated, while the state generator is discarded. Therefore, we can utilize any \ac{RL} algorithm to update the dialogue policy, including both on-policy and off-policy methods. Additionally, since the reward function is pre-trained in an offline manner, we can first infer common information contained in high-quality human-generated dialogues by distinguishing human-generated dialogues from machine-generated ones, and then make full use of the learned information to guide the dialogue policy learning in a new domain in the style of transfer learning.

To summarize, our contributions are:
\begin{itemize}[nosep,leftmargin=*]
\item A reward learning method that is applicable to off-policy \ac{RL} methods in dialogue training.
\item A reward learning method that can alleviate the problem of local optima for adversarial dialogue training.
\item A reward function that can transfer knowledge learned in existing domains to a new dialogue domain.
\end{itemize}

\section{Related Work}
\label{sec:rel_work}
\ac{RL} methods~\citep{peng2017composite, lipton2018bbq,li2017end,Su2018D3Q,dhingra2016towards,williams2017hybrid,li2018dialogue}, have been widely utilized to train a dialogue agent by interacting with users. 
The reward used to update the dialogue policy is usually from a reward function predefined with domain knowledge and it could become very complex, e.g., in the case of multi-domain dialogue scenarios. To provide the dialogue policy with a high quality reward signal, \citet{peng2018adversarial} proposed to make use of the adversarial loss as an extra critic in addition to shape the main reward function. Inspired by the success of adversarial learning in other research fields, \citet{liu2018adversarial} learns the reward function directly from dialogue samples by alternatively updating the dialogue policy and the reward function. The reward function is a discriminator aiming to assign a high value to real human dialogues and a low value to dialogues generated by the current dialogue policy. In contrast, the dialogue policy attempts to achieve higher reward from the discriminator given the generated dialogue. Following this solution, \citet{takanobu2019guided} replaces the discriminator with a reward function a reward function that acts at the dialogue action level, which takes as input the dialogue state, system action, and next dialogue state and returns the reward for the given dialogue action. 

The key distinction of our work compared to previous efforts is being able to train dialogue agents with both: (1)~off-policy methods in adversarial learning settings; (2)~the on-policy based approaches while avoiding potential training issues, such as mode collapse and local optimum. We propose to train (1)~reward model and (2)~dialogue policy \emph{consecutively}, rather than \emph{alternatively} as suggested in ~\cite{liu2018adversarial,takanobu2019guided}. 
To train the reward model, we introduce an auxiliary generator that is used to explore potential dialogue situations. The advantage of our setup is the transfer from SeqGAN~\cite{yu2017seqgan} to a vanilla GAN~\cite{goodfellow2014generative}. In SeqGAN setup, the policy gradient method is essential to deliver the update signal from the discriminator to the dialogue agent. In contrast, in the vanilla GAN, the discriminator can directly backpropogate the update signal to the generator. 
%
%
Once we restore a high-quality reward model, we update the dialogue agent using common \ac{RL} methods, including both on-policy and off-policy.

\section{Learning Reward Functions}
In this section, we introduce our method to learn reward functions with an auxiliary generator.
\subsection{Dialogue State Tracker}
We reuse the rule-based ConvLab dialogue state tracker~\cite{lee2019convlab} to keep track of the information emerging in the interactions,  including the informable slots that show the constraints given by users and requestable slots that indicates what users request. A belief vector is maintained and updated for each slot in every domain.  \\ 
\noindent
\textbf{Dialogue State} 
The collected information from the dialogue state tracker is used to form a structured state representation $state_t$ at every time step $t$. The final representation is formed by (1)~the embedded results of returned entities for a query, (2)~the availability of the booking option with respect to a given domain, (3)~the state of informable slots, (4)~the state of requestable slot, (5)~the last user action, and (6)~the repeated times of the last user action.
The final state representation $S$ is an binary vector with 392 dimensions.\\
\noindent
\textbf{Dialogue Action}
Each atomic action is a concatenation of domain name, action type and slot name, e.g., \textit{Attraction\_Inform\_Address}, \textit{Hotel\_Request\_Internet}. Since in the real scenarios, the response from a human or a dialogue agent can cover combination of atomic actions, we extract the most frequently used dialogue actions from the human-human dialogue collections to form the final action space -- $A$. For example, \textit{[Attraction\_Inform\_Address, Hotel\_Request\_Internet]} is regarded as a new action that the policy agent can execute. The final size of $A$ is $300$. We utilize one-hot embeddings to represent the actions.

\subsection{Exploring Dialogue Scenarios with an Auxiliary Generator}
We aim to train a reward function that has the ability to distinguish high-quality dialogues from unreasonable and inappropriate ones. To generate negative samples, we use an auxiliary generator \textit{Gen} to explore the possible dialogue scenarios that could happen in real life. The dialogue scenario at time $t$ is a pair of a dialogue state $s_t$ and the corresponding system action $a_t$. The dialogue state-action pairs generated by \textit{Gen} are fed to the reward model as negative samples. During reward training, the reward function can benefit from the rich and high-quality negative instances generated by the advanced generator $Gen$ to improve the discriminability. Next, we will explain how states and actions are simulated, and our setup for adversarial leaning.

\subsubsection{Action Simulation}
To simulate the dialogue actions, we use a \ac{MLP} as the action generator $\textit{Gen}_a$ followed by a Gumbel-Softmax function with $300$ dimensions, where each dimension corresponds to a specific action from the defined $A$. The Gumbel-Max trick~\citep{gumbel1954statistical} is commonly used to draw a sample $u$ from a categorical distribution with class probabilities $p$:
\begin{equation}
u = one\_hot (\argmax_i [g_i + \log p_i])
\end{equation}
where $g_i$ is independently sampled from Gumbel $(0,1)$. Since the $\argmax$ operation is not differentiable, no gradient can be backpropagated through $u$. Instead, we employ the soft-argmax approximation~\citep{jang2016categorical} as a continuous and differentiable approximation to $\argmax$ and to generate the $k$-dimensional sample vector $y$ following:
\begin{equation}
   y_i = \frac{\exp((\log(p_i) + g_i)/\tau)}{\sum_{j=1}^k \exp((\log(p_j) + g_j)/\tau)} 
\label{eq:gen_fake}
\end{equation}
for $i=1,\ldots, k$. 
When the temperature $\tau \rightarrow 0$, the $\argmax$ operation is exactly recovered but the gradient will vanish. In contrast, when $\tau$ goes up, the Gumbel-Softmax samples are getting similar to samples from a uniform distribution over $k$ categories. In practice, $\tau$ should be selected to balance the approximation bias and the magnitude of gradient variance. In our work, $p$ corresponding to the output distribution of generator $\textit{Gen}_a$ and $k$ equals to the action dimension $300$.

\subsubsection{State Simulation}
Compared to the GANs scenarios in computer vision, the output of the generator in our setting is a discrete vector which makes it challenging to backpropogate the loss from discriminator to the generator directly. To address this problem, we propose to project the discrete representation $x$ in the expert demonstrations to a continuous space with an encoder \textit{Enc} from a pre-trained variational autoencoder~\citep{kingma2013auto}. We assuming the human-human dialogue state $s$ is generated by a latent variable $z_{vae}$ via the decoder \textit{Dec} $p(s|z_{vae};\psi)$. Then we can regard the variable $z_{vae}$ as a desired representation in a continuous space. Given a human-generated state $s$, the VAE utilizes a conditional probabilistic encoder $Enc$ to infer $z_{vae}$ as follows:
\begin{equation}
z_{vae} \sim \textit{Enc}(s) = q_\omega(z_{vae}|s),
\end{equation}
where $\omega$ and $\psi$ are the variational parameters encoder and decoder respectively. The optimization objective is given as:
\begin{equation}
\begin{split}
\mbox{}\hspace*{-2mm}L_{vae}(\omega, \psi) &= \underbrace{\mathbb{E}_{z_{vae}\sim q_{\omega}(z_{vae}|s)}[\log p_\psi(s | z_{vae})]}_{(1)} \\
&\quad +  \underbrace{\textit{KL}(q_\omega(z_{vae}|s)||p(z_{vae}))}_{(2)},
\end{split}
\hspace*{-1mm}\mbox{}
\end{equation}
where (1) is responsible for encouraging the decoder parameterized with $\psi$ to learn to reconstruct the input $x$; (2) is the KL-divergence between the encoder distribution $q_\omega(z_{vae}|s;\omega)$ and a standard Gaussian distribution $p(z_{vae})=N(0,I)$.

\begin{figure}[t]
\centering
   \includegraphics[clip, width=1.0\columnwidth]{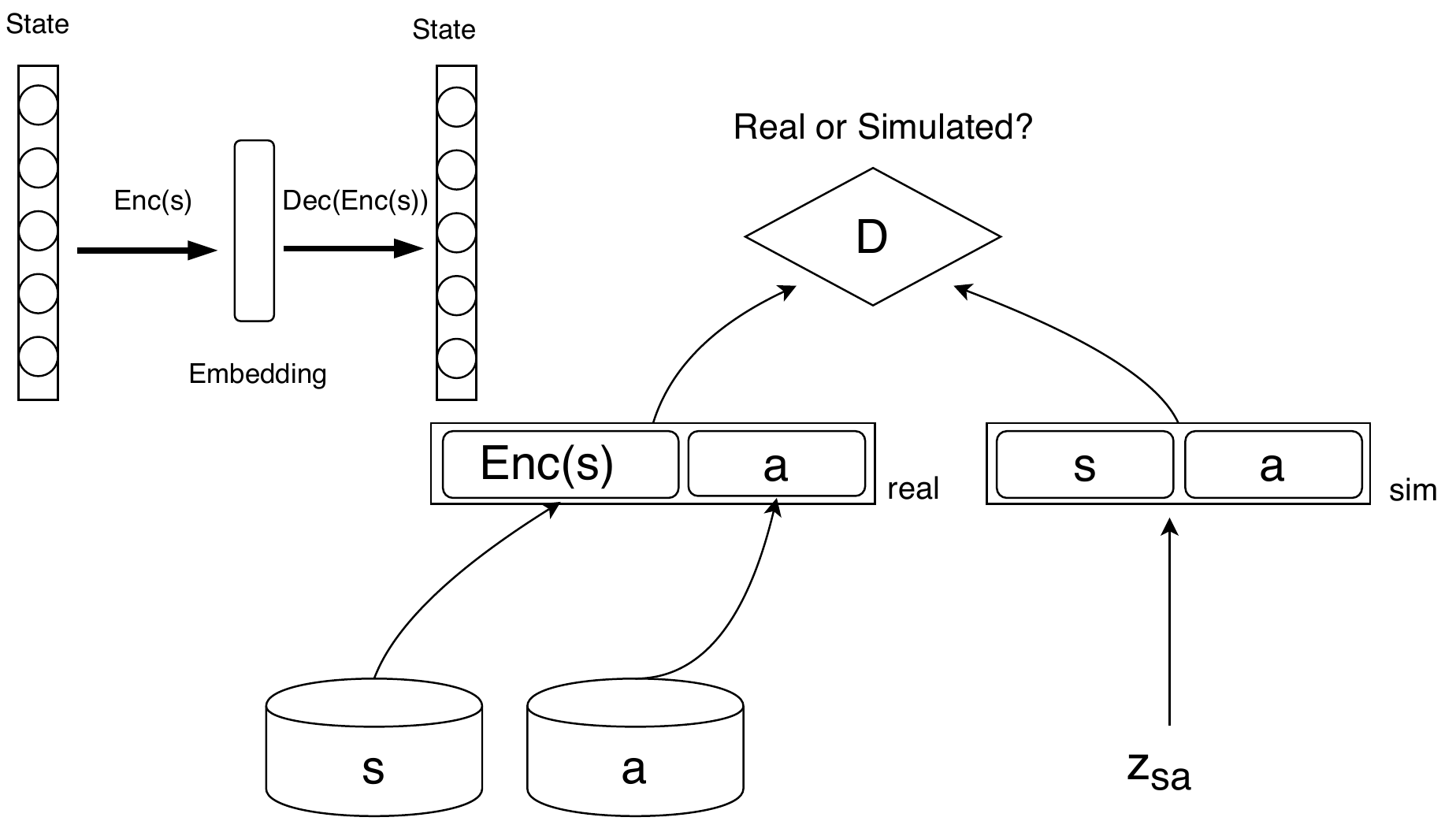}
   \caption{The architecture to simulate state-action representations with a variational autoencoder. $z_{sa}$ is the sampled Gaussian noise.}
   \label{fig:vae_gan}
\end{figure}

The benefit of projecting the state representations to a new space is directly simulating the dialogue states in the continuous space $S_{\textit{embed}}$ similar to generating realistic images in computer vision. Besides, similar dialogue states are embedded into close latent representations in the continuous space to improve the generalizability. Figure~\ref{fig:vae_gan} shows the overall process of learning the state projecting function $\textit{Enc}_{\omega}(s)$ given dialogue states from real human-human dialogues. We use $s_{\textit{real}}$ to denote the continuous representation of real state $s$ while $s_{\textit{sim}}$ for the simulated one. 

\subsubsection{Adversarial Training}
We can approximate the real state-action distribution in a differentiable setup (1)~by applying Gumbel-Softmax to simulate actions $a_{\textit{sim}}$; and (2)~by directly generating simulated states $s_{\textit{sim}}$ in the continuous space $S_{\textit{embed}}$.
The auxiliary generator $\textit{Gen}_\theta$ to simulate $s_{\textit{sim}}$ and $a_{\textit{sim}}$ has following components:
\begin{equation}
\begin{split}
h & = \textit{MLP}_1(z_{sa})\\
a_{\textit{sim}} & = f_{\textit{Gumbel}}(\textit{MLP}_2(h)) \\
s_{\textit{sim}} & = \textit{MLP}_3(h)\\
(s,a)_{\textit{sim}} & = s_{\textit{sim}} \oplus a_{\textit{sim}} 
\end{split}
\end{equation}

\noindent
where $\theta$ denotes all the parameters in the generator and $\oplus$ is the concatenation operation. During the adversarial training process, the generator $\textit{Gen}_\theta$ takes noise $z_{sa}$ as input and outputs a sample $(s,a)_{\textit{sim}}$ and it aims to get higher reward signal from the discriminator $D_\phi$. The training loss for the generator $\textit{Gen}_\theta$ can be given as:
\begin{equation}
    L_{G}(\theta) = - \mathbb{E}_{(s,a)_{\textit{sim}} \sim \textit{Gen}_\theta}(R_{\phi}((s,a)_{\textit{sim}}),
\end{equation}
where $R_\phi((s,a)_{\textit{sim}}) = - \log(1 - D_\phi((s,a)_{\textit{sim}})$ and $D_\phi$ denotes the discriminator measuring the reality of generated state-action pairs $(s,a)_{\textit{sim}}$.

The discriminator $D_\phi$ in this work is a \ac{MLP} that takes as input the state-action pair $(s,a)$ and outputs the probability $D(s,a)$ that the sample is from the real data distribution. Since the discriminator's goal is to assign higher probability to the real data while lower scores to simulated data, the objective can be given as the average log probability it assigns to the correct classification. Given an equal mixture of real data samples and generated samples from the generator $\textit{Gen}_\theta$, the loss function for the discriminator $D_\phi$ is:
\begin{align}
L_{D}&(\phi)={} \nonumber\\
& \mathbb{E}_{((s,a)_{\textit{sim}}) \sim \textit{Gen}_\theta}(\log(1 - D_\phi((s,a)_{\textit{sim}}))) \nonumber\\ 
& - \mathbb{E}_{(s,a) \sim \textit{data}}(D_\phi(\textit{Enc}_\omega(s),a)_{\textit{real}})).
\label{eq:gan_loss}
\end{align}
After the adversarial training is finished, we will keep the discriminator $D_\phi$ as the reward function for future dialog agent training while the generator $\textit{Gen}_\theta$ will be discarded.

Next, we discuss a suitable experimental environment for validating the presented method.

\section{Experiemntal Setup}
\label{exp_setup}

\subsection{Dataset and Training Environment}
\label{exp_data}
\textbf{MultiWOZ}~\citep{budzianowski2018multiwoz} is a multi-domain dialogue dataset spanning $7$ distinct domains\footnote{Attraction, Hospital, Police, Hotel, Restaurant, Taxi, Train}, and $10,438$ dialogues. The main scenario in this dataset is that a dialogue agent is trying to satisfy the demand from tourists such as booking a restaurant or recommending a hotel with specific requirements. 
The average number of turns is 8.93 and 15.39 for single and multi-domain dialogues, respectively.\\
\noindent
\textbf{ConvLab}~\citep{lee2019convlab} is an open-source multi-domain end-to-end dialogue system platform offering the annotated MultiWOZ dataset and associated pre-trained reference models. We reuse the rule-based dialogue state tracker from ConvLab to track the information that emerges during interactions between users and the dialogue agent. Besides, an agenda-based~\citep{schatzmann2007agenda} user simulator is embedded and used for multi-domain dialogue scenarios.\\
\noindent
\textbf{Evaluation metrics} Before a conversation starts, a user goal will be randomly sampled. The user goal consists of two parts: (1)~the constraints on different domain slots or booking requirements, e.g., \textit{Restaurant\_Inform\_Food$=$Thai}; (2)~the slot values that show what the user is looking for, e.g., \textit{Restaurant\_Request\_phone$=$?}. We The task is completed successfully, if a dialogue agent has provided all the requested information and made a booking according to the requirements. We use \emph{average turn} and \emph{success rate} to evaluate the efficiency and level of task completion of dialogue agents. 

\subsection{Architecture and Training Details}
\textbf{Variational AutoEncoder} The encoder is a two-layer \ac{MLP} that takes the discrete state representation ($392$ dimensions) as input and outputs two intermediate embeddings (64 dimensions) corresponding to the mean and the variance, respectively. For inference, we regard the mean $\mu$ as the embedded representation for a given state input $s$. \\
\noindent
\textbf{Auxiliary Generator}
The auxiliary generator takes randomly sampled Gaussian noise as input and outputs a continuous state representation and a one-hot action embedding. The input noise is fed to a one-layer MLP first followed by the state generator $Gen_s$ and action generator $Gen_a$. $Gen_s$ is implemented with a two-layer \ac{MLP} which output is the simulated state representation ($64$ dimensions) corresponding to the input noise. The main component of $Gen_a$ is a two-layer \ac{MLP} followed by a Gumbel-Softmax function. The output of the Gumbel-Softmax function is an one-hot representation ($300$ dimensions). Specifically, we implemented the `Straight-Through' Gumbel-Softmax Estimator~\citep{jang2016categorical} and the temperature for the function is set to $0.8$.\\
\noindent
\textbf{Discriminator}
The discriminator is a three-layer \ac{MLP} that takes as input the concatenation of latent state representation ($64$ dimensions) and one-hot encoding of the action ($300$ dimensions). During adversarial training, the real samples come from the real human dialogues in the training set while the simulated samples have three different sources. 
The main source is the output of the auxiliary generator introduced above. 
The second one is a random sample of state-action pairs from the training set where the action in each pair is replaced with a different one to build a simulated state-action pair. As a third source, we keep a history buffer with size $10k$ to record the simulated state-action pairs from the generator, where the state-action pairs are replaced randomly by the newly generated pairs from the generator. To strengthen the reward, we incorporate the human feedback $r_{\textit{Human}}$ into the pre-trained reward function.  As the final reward function to train the dialogue agent we use the mixed reward $r_\textit{GAN-VAE}=r_\textit{Human} + \log(D(s,a))$. 

\subsection{Reinforcement Learning Methods}
In this work, we validate our pre-trained reward using two different types of \ac{RL} methods: \ac{DQN}~\citep{mnih2015human}, which is an off-policy \ac{RL} algorithm, and \ac{PPO}~\citep{schulman2017proximal}, which is a policy-gradient-based \ac{RL} method. 
To speed up the training speed, we extend the vanilla \ac{DQN} to WDQN, where the real dialogue state-action pairs from the training set are used to warm up the dialogue policy at the very beginning and then gradually removed from the training buffer.
We implemented the \ac{DQN} and \ac{PPO} algorithms according to the ConvLab \ac{RL} module\footnote{The code of our work: \url{https://github.com/cszmli/dp-without-adv}}.

\subsection{Baselines}
\label{exp:baselines}

The handcrafted reward function $r_\textit{Human}$ is defined at the conversation level as follow: if the dialogue agent successfully accomplish the task within $T$ turns, it will receive $T * 2$ as reward; otherwise, it will receive $-T$ as penalty. $T$ is the maximum number of dialogue turns. $T$ is set $40$ for experimentation. Furthermore, the dialogue agent will receive $-1$ as intermediate reward during the dialogue to encourage shorter interactions. 

In terms of \ac{DQN}-based methods, we have \textit{DQN(Human)} trained with $r_\textit{Human}$ and \textit{DQN(GAN-VAE)} trained with $r_\textit{GAN-VAE}$. We also develop a variant \textit{DQN(GAN-AE)} by replacing the variational autoencoder in \textit{DQN(GAN-VAE)} with an vanilla autoencoder.  
With respect to WDQN, we provide three different dialogue agents trained with reward functions from \textit{Human}, \textit{GAN-AE}, and \textit{GAN-VAE}.

In terms of \ac{PPO}-based methods, we implemented \ac{GAIL}~\cite{gan_imitation} and \ac{AIRL}~\citep{takanobu2019guided}. In \textit{GAIL}, the reward is provided with a discriminator where its parameter will be updated during the adversarial training process. \textit{AIRL} is an adversarial learning method as well. The difference is that the discriminator in \textit{GAIL} is replaced with a reward function that acts at the action level, which takes as input the dialogue state, system action, and the next state and returns the reward for the given dialogue action. For a fair comparison, both the \textit{GAIL} discriminator and the \textit{AIRL} reward model have been pre-trained. We also utilize teacher-forcing~\citep{bengio2015scheduled} for human dialogues to stabilize the adversarial training process. 

Next, we report the average performance by running the same method $8$ times with different random seeds.

\section{Experimental Results}

\subsection{Results with DQN-based agents}
\label{sec:results_dqn}
\begin{figure}[t]
\centering
   \includegraphics[clip, width=1.0\columnwidth]{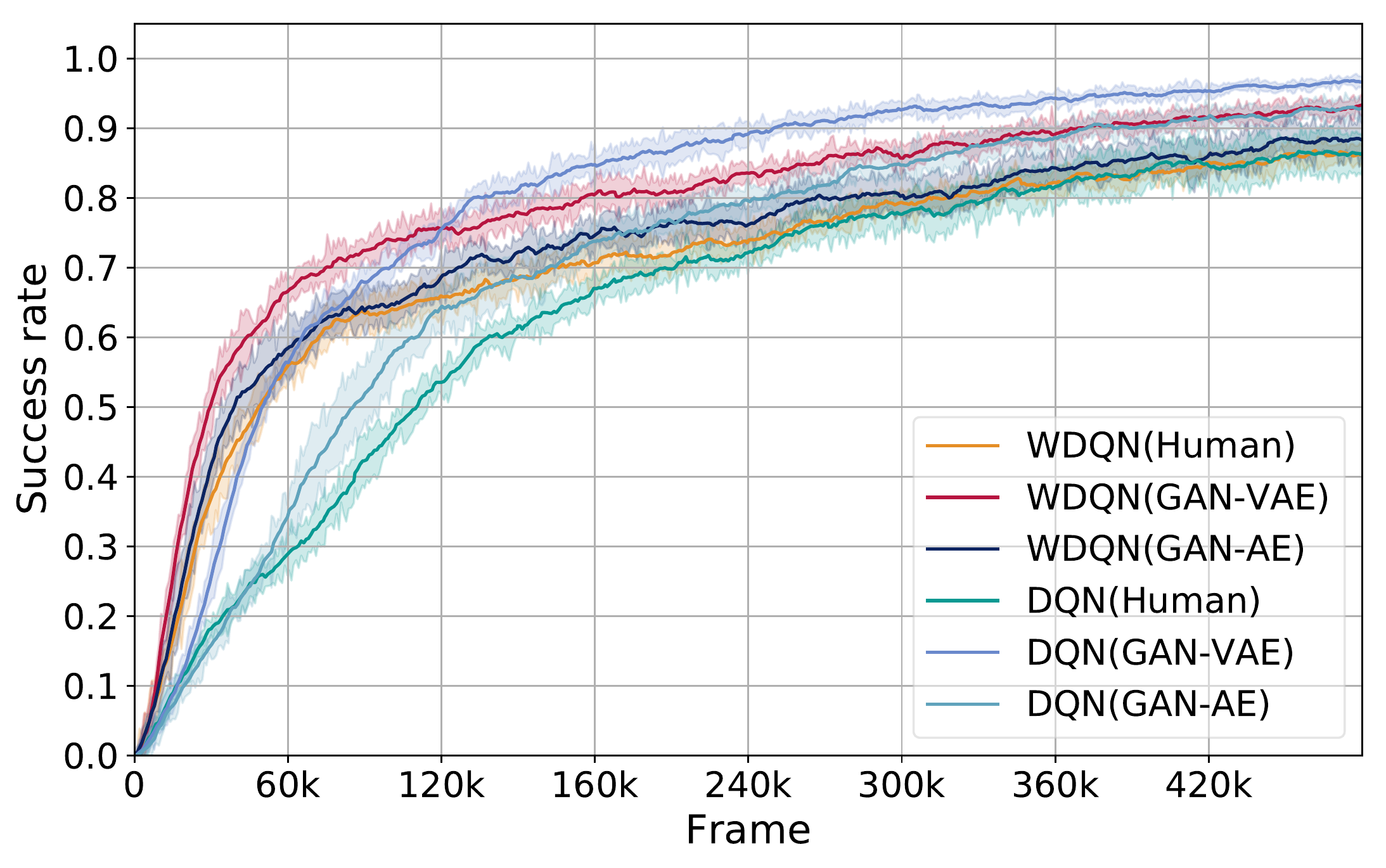}
   \caption{The learning process of DQN-based dialogue agents with different reward functions.}
   \label{fig:learning_curve}
\end{figure}

Figure~\ref{fig:learning_curve} plots the results of DQN-based methods with different reward functions but the same user simulator. The dialogue policy trained with \textit{GAN-VAE} shows the best performance in terms of convergence speed and success rate. In comparison with \textit{GAN-VAE} and \textit{GAN-AE}, the updating signal from the handcrafted reward function $r_\textit{Human}$ can still guide the dialogue policy to a reasonable performance but with a slower speed. This suggests that denser reward signals could speed up the dialogue policy training. Moreover, the policy with $r_\textit{Human}$ converges to a lower success rate compare to \textit{GAN-VAE} and \textit{GAN-AE}. It suggests that, to some extent, the pre-trained reward functions have mastered the underlying information to measure the quality of given state-action pairs. The knowledge that the reward function learned during the adversarial learning step could be generalized to unseen dialogue states and actions to avoid a potential local optimum. In contrast, the dialogue agent \textit{DQN(Human)} only relies on the final reward signal from the simulator at the end of dialogue, which cannot provide enough guidance to the ongoing turns during conversations. This could be the reason why \textit{DQN(Human)} shows lower success rate compare to \textit{DQN(GAN-VAE)} and \textit{DQN(GAN-AE)}. The representation quality of the learned state embeddings leads to higher \textit{GAN-VAE} performance over \textit{GAN-AE}, because \textit{VAE} generalizes better thereby bringing more benefits to the reward functions.   

Examining closer WDQN agents, we can see all three methods achieve their inflection points after the first $30k$ frames.
Comparing \textit{DQN(Human)} and \textit{WDQN(Human)}, we found the real human-human generated dialogue pairs from training set do alleviate the problem of sparse reward provided by $r_\textit{Human}$ at the start stage of policy training. 
Similar results could be observed from agents trained with the pre-trained reward function $r_\textit{GAN-VAE}$. 
After $24k$ frames, the \textit{WDQN(Human)} curve coincides in position with \textit{DQN(Human)} and they converge to the same point in the end. The faster convergence speed on \textit{WDQN(Human)} did not bring a higher success rate because the dialogue policy still has no access to precise intermediate reward signals for the ongoing dialogue turns.

\begin{table}[ht]
  \centering
  \resizebox{\columnwidth}{!}{
\begin{tabular}{l*{2}{c}}
\toprule
\textbf{Dialogue agent} & \textbf{Success Rate} & \textbf{Average Turn} \\
\midrule
$\text{WDQN}_{keep}$(Human)  & 0.741 & 19.144  \\
$\text{WDQN}_{keep}$(GAN-AE)  & 0.879 &15.118     \\
\midrule
WDQN(Human)   & 0.906  & 13.580   \\
WDQN(GAN-AE)  &0.911 & 13.298    \\
WDQN(GAN-VAE)  & 0.937  & 12.260   \\
\midrule
DQN(Human)  & 0.870  & 14.960    \\
DQN(GAN-AE) & 0.953 & 12.300     \\
DQN(GAN-VAE) & \textbf{0.985} & \textbf{11.040}     \\
\bottomrule
\end{tabular} 
}
    \vspace*{0.5\baselineskip}
\caption{The final performance of DQN-based dialogue agents with different reward functions.}
\label{Table:results}
\end{table}
Table~\ref{Table:results} reports the final performance of different dialogue agents during test time. All the agents have been trained with $500k$ frames and we save and evaluate the model that has the best performance during the training stage. Interestingly, \textit{DQN(GAN-VAE)} outperforms \textit{WDQN(GAN-VAE)} while \textit{WDQN(Human)} beats \textit{DQN(Human)}. The warming-up stage in \textit{WDQN(GAN-VAE)} does improve the training speed but it resulted in a lower final success rate. The potential reason is that the real human-human dialogue can bring a strong update signal at the beginning of the training process but at the same time limits the exploration ability of the agent. 
To verify this finding, we designed two more WDQN agents: \textit{WDQN$_{keep}$(Human)} and \textit{WDQN$_{keep}$(GAN-AE)},
which keep expert dialogues examples during the entire training phase, rather than removing them gradually.
Their performance is shown in Table~\ref{Table:results}. As to agents trained with $r_\textit{Human}$, there is a huge performance gap, \textit{WDQN(Human)} outperforms \textit{WDQN$_{keep}$(Human)} almost by $15\%$. The difference in the performance of \textit{WDQN$_{keep}$(GAN-AE)} and \textit{WDQN(GAN-AE)} is significantly smaller because the pre-trained reward function brings more precise and consistent update signals that are explored and disclosed during the adversarial training step. 

\begin{figure}[ht]
\centering
   \includegraphics[clip, width=1.0\columnwidth]{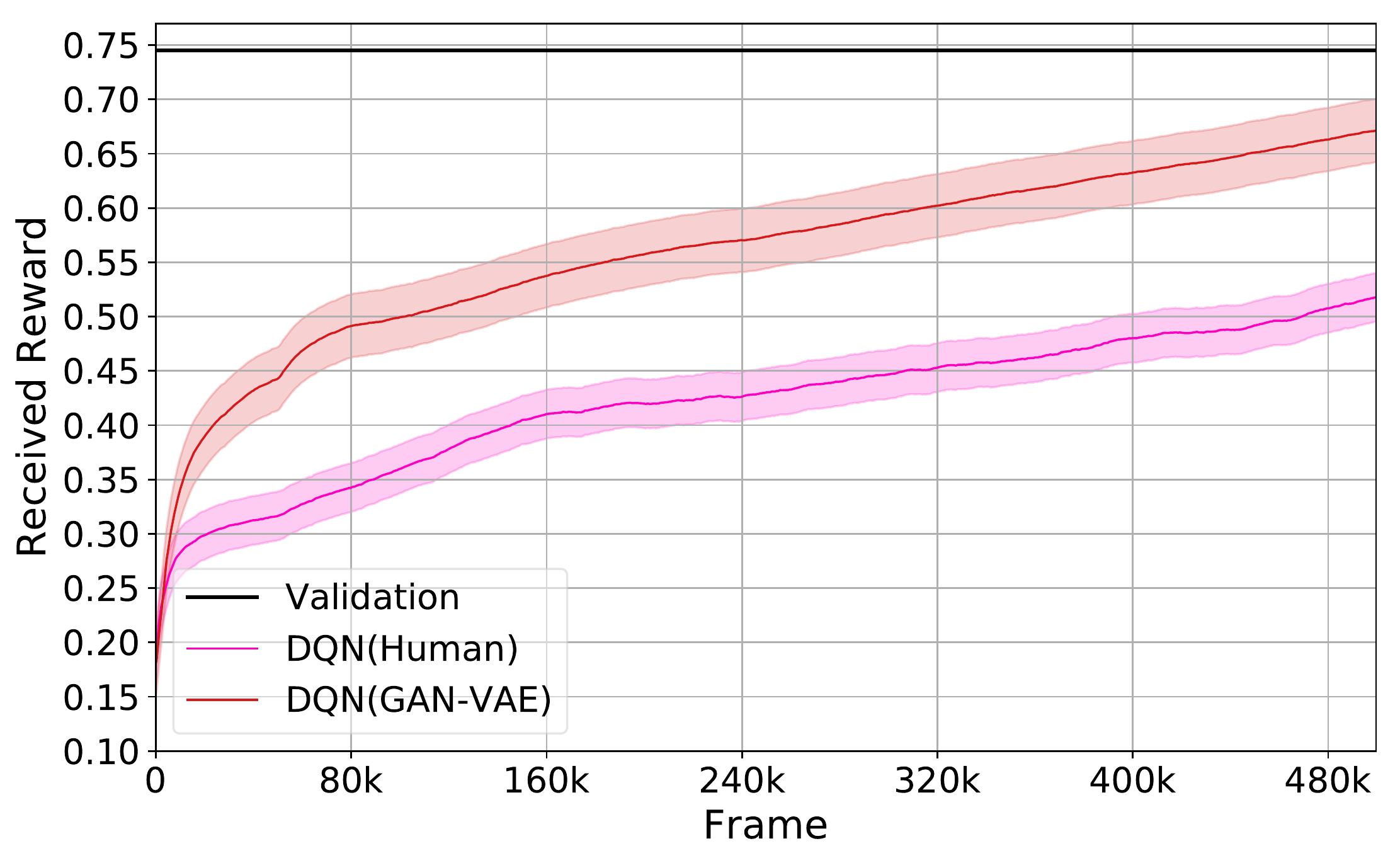}
  \caption{The reward returned by the pre-trained reward function during dialogue policy training.}
   \label{fig:dqn_reward_curve}
\end{figure} 
Figure~\ref{fig:dqn_reward_curve} shows curves presenting the reward changes during the \ac{RL} training. The curve \textit{Validation} denotes the average reward received based on the real human-human dialogues, which can be regarded as the human performance evaluated by the pre-train reward function $r_\textit{GAN-VAE}$ and it is $\sim 0.74$.\footnote{Ideally, the reward on human dialogues should be equals to 0.5 because the discriminator is not able to distinguish the simulated dialogues from real human-human ones after generator and discriminator converge according to Eq.~\ref{eq:gan_loss}.} For \textit{DQN(Human)} and \textit{DQN(GAN-VAE)} training, we feed generated in real-time dialogue batches to reward function $r_\textit{GAN-VAE}$. We can see that both approaches are getting a high reward, but \textit{DQN(GAN-VAE)} is growing faster, because $r_\textit{GAN-VAE}$ is used for the training of \textit{DQN(GAN-VAE)}. That is a promising finding since we can suggest that a well-trained reward function can be utilized not only to guide the dialogue policy training but also to judge the quality of different agents.

\subsection{Results with PPO-based agents}

As for \textit{\ac{GAIL}} and \text{\ac{AIRL}}, the reward functions are updated on the fly, and therefore we can only employ policy gradient-based \ac{RL} algorithms. 
We use \ac{PPO} algorithms to train the dialogue agent with different reward functions. Before initiating training, we first warm-up all the dialogue agents with human dialogues via imitation learning. As a result, the warmed-up agents share similar success rates which is $\sim 33\%$. We also pre-train discriminators in \textit{GAIL} and reward models in \textit{\ac{AIRL}} utilizing positive examples from the training set and negative examples from the pre-trained dialogue agents.

\begin{figure}[ht]
\centering
   \includegraphics[clip, width=1.0\columnwidth]{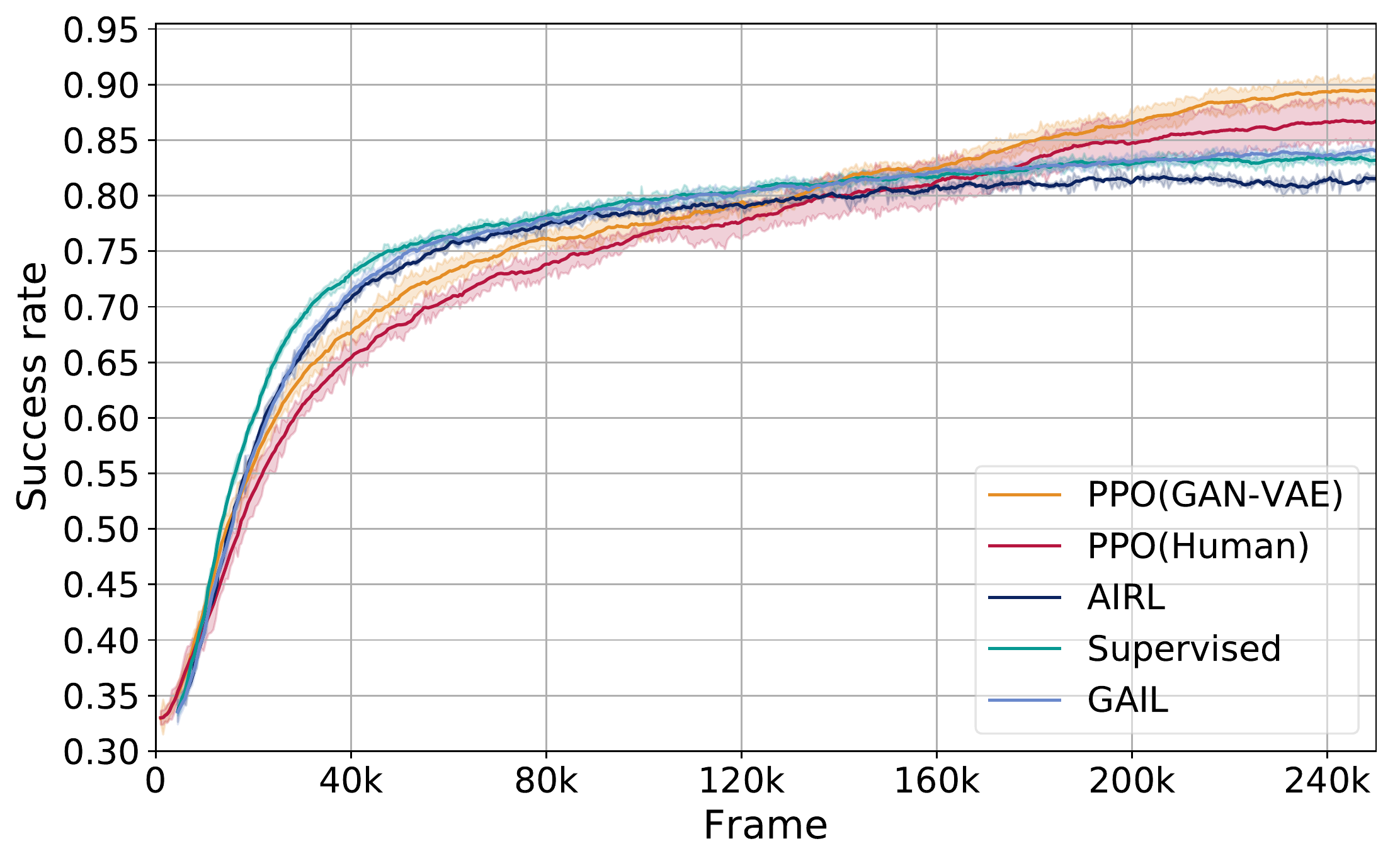}
   \caption{The learning process of \ac{PPO}-based dialogue agents with different reward functions.}
   \label{fig:ppo_learning_curve}
\end{figure} 

Figure~\ref{fig:ppo_learning_curve} demonstrates that in terms of success rate \textit{GAIL} and \textit{AIRL} rise faster than \textit{PPO(GAN-VAE)} and \textit{PPO(Human)} during first $120k$ frames. Then both methods flattened and converged at $\sim 81\%$.
It is important to note, that we utilize teacher-forcing in the adversarial step by feeding human-human dialogues to the agents every several frames while training \textit{\ac{GAIL}} and \textit{\ac{AIRL}}. Due to the large task action space, it is nearly impossible to successfully train a high-quality dialogue agent without teaching-forcing steps in adversarial learning methods. %
The agent called \textit{supervised} represents the setup where we discard the training signals from the discriminators or the reward models in \textit{GAIL} and \textit{AIRL} and only train the policy network using teacher-forcing with the same frequency. We can observe that the adversarial training signal in \textit{GAIL} and \textit{AIRL}  degenerates the performance of supervised learning methods. 
\noindent
\subsubsection{Discussion}
We explored various parameters for \textit{\ac{GAIL}} and \textit{\ac{AIRL}} setups, unfortunately unsuccessful. The potential reason is ConvLab has $300$ actions, and it is intractable for a dialogue agent to explore the action space relying only on the sparse positive reward signals which can easily lead to a local optimum. \citet{takanobu2019guided} successfully applied \textit{AIRL} to learn dialogue policy, but the considered size of action space was only half compared to our setup. More importantly,~\citet{takanobu2019guided} formulated dialogue policy learning as a multi-label classification task where it is easier to achieve a higher success rate by selecting as many actions as possible in one turn. Moreover, DQN-based \ac{RL} algorithms are not applicable in their setup. 
In comparison, our agent \textit{PPO(GAN-VAE)} can achieve higher performance in the more commonly used setup. 
Comparing \textit{PPO(GAN-VAE)} and \textit{PPO(Human)}, we can verify our claim that the dialogue agent benefits from the pre-trained reward function $r_\textit{GAN-VAE}$. As shown in Figure~\ref{fig:learning_curve} and  Figure~\ref{fig:ppo_learning_curve}, the agents trained using the hand-crafted reward function, such as \textit{DQN(Human)} and \textit{PPO(Human)}, share a similar final performance $ \sim 87\%$. Another important finding the DQN-based agents benefit more compared to the PPO-based ones from incorporating the reward signals from the same reward function $r_\textit{GAN-VAE}$.

\begin{figure}[t]
\centering
   \includegraphics[clip, width=1.0\columnwidth]{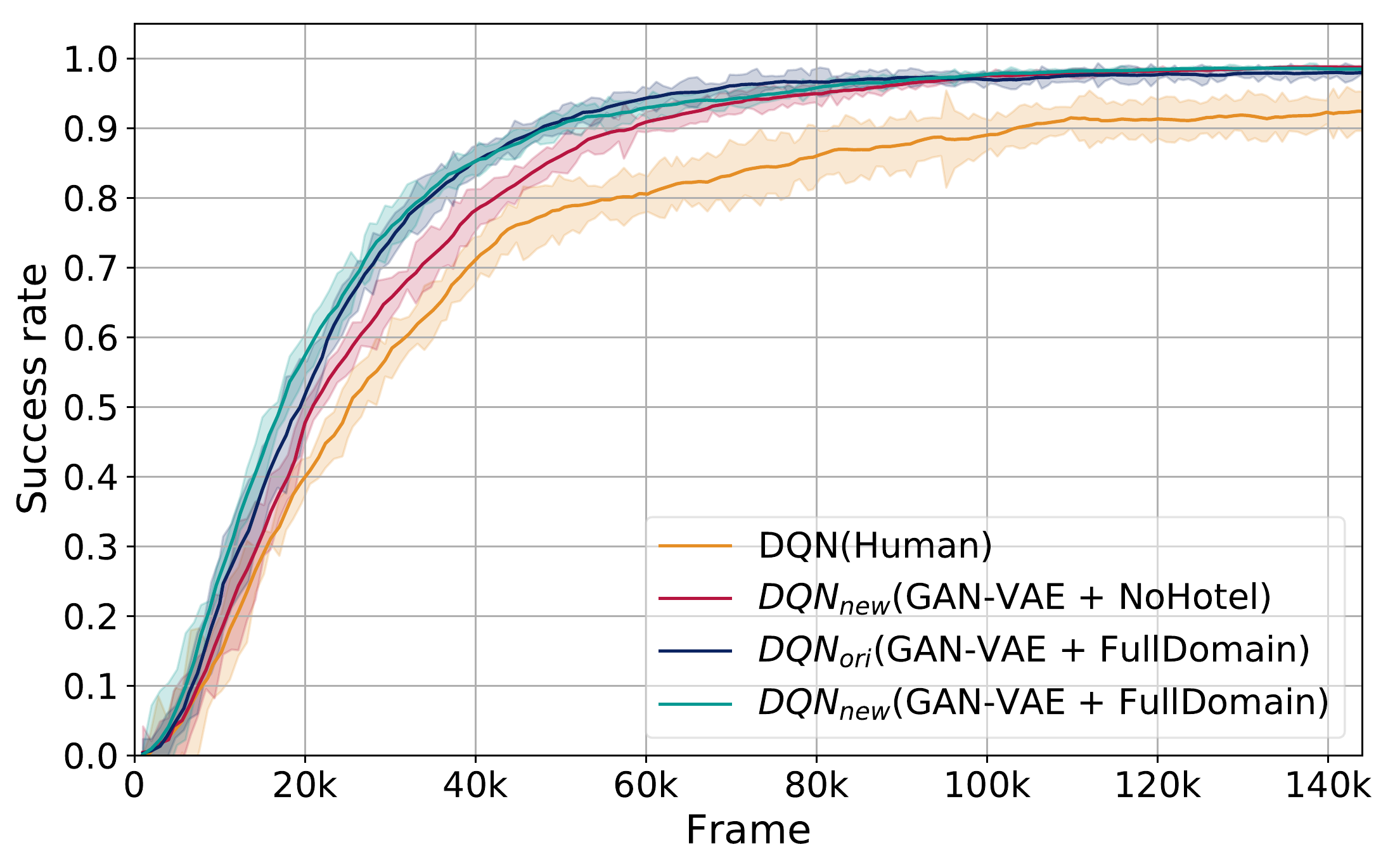}
   \caption{The learning process of dialogue agents in different domains.}
   \label{fig:dqn_nohotel_learning_curve}
\end{figure} 

\subsection{Transfer learning with pre-trained reward function}
To define the action space, we utilize  $300$ the most frequent actions from the MultiWoz dataset and use one-hot embedding to represent them. 
As shown in Figure~\ref{fig:vae_gan}, the action and the state representations are concatenated to form a specific state-action pair. 
This approach ignores the relations between different actions. For example, \textit{Restaurant\_Inform\_Price} and \textit{Restaurant\_Request\_People} should be close for the same conversation since they happen to be  in the same domain. However, even for different domains, connections between actions are possible, e.g. \textit{Inform\_Price} and \textit{Request\_People} can also happen in the \textit{Hotel} domain, corresponding to actions \textit{Hotel\_Inform\_Price} and \textit{Hotel\_Request\_People}. 
We ask ourselves if we can transfer the knowledge learned in existing domains to a new domain, which we have never seen before via the pre-trained reward function. 
To answer this question, we first reformulate the action representation as a concatenation of three different segments: \textit{Onehot(Domain)}, \textit{Onehot(Diact)}, \textit{Onehot(Slot)}. Following this approach, actions containing similar information will be linked through the corresponding segments in their representation. Utilizing this formulation, we retrained our reward function in selected domains and incorporate it into the training of a dialogue agent in a new unseen domain. Concretely,  we train the reward function based on the following domains: \textit{Restaurant, Bus, Attraction, and Train}. As a testing domain, we pick  \textit{Hotel} since it has the most slot types and some of them are unique, such as \textit{Internet, Parking, Stars}. \textit{DQN}$_{\textit{ori}}$ in Figure~\ref{fig:dqn_nohotel_learning_curve} corresponds to the dialogue agent trained with all domains and the action is represented with a single one-hot embedding. By replacing the action representation in \textit{DQN}$_{\textit{ori}}$ with the new action formulation we get agent -- \textit{DQN}$_{\textit{new}}$. Based on the obtained results, we can conclude \textit{DQN$_{\textit{new}}$(GAN-VAE + NoHotel)} benefits from the reward function trained in different domains and it outperforms \textit{DQN(Human)}. As expected, the agents \textit{DQN$_{\textit{new}}$(GAN-VAE + FullDomain)} and \textit{DQN$_{\textit{ori}}$(GAN-VAE + FullDomain)}, which are trained using reward from all domains, have better performance compared to \textit{DQN$_{\textit{new}}$(GAN-VAE + NoHotel)}.

\section{Conclusion}
In this work, we have proposed a guided dialogue policy training method without using adversarial training in the loop. 
First, we trained the reward model with an auxiliary generator. Then the trained reward model was incorporated into a common reinforcement learning method to guide training of a high-quality dialogue agent. By conducting extensive experimentation, we demonstrated that the proposed methods achieve remarkable performance, in terms of task success, as well as the potential to transfer knowledge from previously utilized task domains to new ones.

\bibliographystyle{acl_natbib}
\bibliography{bibliography}
\end{document}